\pdfoutput=1

\documentclass[11pt]{article}

\usepackage{acl}

\usepackage{times}
\usepackage{latexsym}

\usepackage[T1]{fontenc}

\usepackage[utf8]{inputenc}

\usepackage{microtype}

\usepackage{inconsolata}
\usepackage{booktabs}
\usepackage{amssymb}
\usepackage{pifont}
\usepackage{subcaption}
\usepackage{svg}
\usepackage{multirow}
\usepackage{multicol}
\usepackage{comment}
\usepackage{enumitem}

\title{Can LMs Generalize to Future Data? \\ An Empirical Analysis on Text Summarization}

\author{Chi Seng Cheang$^1$~~~
        Hou Pong Chan$^1\thanks{~~Co-corresponding author.}$~~~
        Derek F. Wong$^{1,2}\footnotemark[1]$~~~
        Xuebo Liu$^3$~~~ \\
        \textbf{Zhaocong Li}$^1$~~~ 
        \textbf{Yanming Sun}$^1$~~~
        \textbf{Shudong Liu}$^1$~~~
        \textbf{Lidia S. Chao}$^1$~~~\\
  $^1$NLP$^2$CT Lab, Department of Computer and Information Science, 
  University of Macau \\
  $^2$Institute of Collaborative Innovation, University of Macau\\
    $^3$Institute of Computing and Intelligence, Harbin Institute of Technology, Shenzhen, China \\
    \texttt{andy.cheang@connect.um.edu.mo, \{hpchan,derekfw,lidiasc\}@um.edu.mo} \\
          \texttt{liuxuebo@hit.edu.cn, nlp2ct.\{zhaocong,yanming,shudong\}@gmail.com} } 

\begin{document}
{\makeatletter\acl@finalcopytrue
  \maketitle
}
\begin{abstract}
Recent pre-trained language models (PLMs) achieve promising results in existing abstractive summarization datasets. 
However, existing summarization benchmarks overlap in time with the standard pre-training corpora and finetuning datasets. Hence, the strong performance of PLMs may rely on the parametric knowledge that is memorized during pre-training and fine-tuning. Moreover, the knowledge memorized by PLMs may quickly become outdated, which affects the generalization performance of PLMs on future data. 
In this work, we propose \textsc{TempoSum}, a novel benchmark that contains data samples from 2010 to 2022, to understand the temporal generalization ability of abstractive summarization models.
Through extensive human evaluation, we show that parametric knowledge stored in summarization models significantly affects the faithfulness of the generated summaries on future data. 
Moreover, existing faithfulness enhancement methods cannot reliably improve the faithfulness of summarization models on future data. 
Finally, we discuss several recommendations to the research community on how to evaluate and improve the temporal generalization capability of text summarization models.\footnote{Our \textsc{TempoSum} benchmark is available at \href{https://github.com/NLP2CT/TempoSum}{https://github.com/NLP2CT/TempoSum}.} 
\end{abstract}

\section{Introduction}

Abstractive summarization aims to generate a concise summary that contains the critical information of the source text while ensuring the generated text is fluent and faithful.  
In recent years, pre-trained language models (PLMs) have accomplished state-of-the-art outcomes across numerous summarization benchmarks~\cite{DBLP:conf/icml/ZhangZSL20, DBLP:conf/acl/LewisLGGMLSZ20}. 
Nonetheless, the test sets within these benchmarks are predominantly comprised of data samples prior to 2019, resulting in a substantial temporal overlap with both the training sets and standard pre-training corpora.
As summarization systems are commonly deployed in practical applications for processing text articles from future time periods, it is crucial to consider their performance on such data.
PLMs that excel on data originating from the same temporal context as the pre-trained corpus may not necessarily exhibit robust generalization capabilities when applied to future datasets. 

\begin{figure}[t!]
  {\fontsize{10}{12}\selectfont 
    \begin{tabular}{ |p{7.2cm}| }
    \hline
      \textbf{News article from CNN in 2015:} \\ Add \textcolor{red}{Vice President Joe Biden} to the list of potential 2016 candidates traveling to the first-in-the-nation caucus state. \textcolor{red}{Biden} will travel on official White House business {next week} {to} Des Moines, {Iowa} where he will deliver a speech on the administration' s economic policies, his office announced Friday ...\\
      \textbf{Generated Summary:} \\\textcolor{red}{Vice President Joe Biden} will travel to Iowa next week ...  \\ 
      \hline
      \hline
      \textbf{News article from CNN in 2021:} \\ \textcolor{red}{President Joe Biden}’s mission to speed up the lackluster rollout of coronavirus vaccines. \textcolor{red}{Biden} signed a slew of {executive orders} this week aimed at accelerating the {distribution} of {vaccines} and tests ... \\
      \textbf{Generated Summary:} \\ \textcolor{red}{\underline{Vice President} Joe Biden} has signed executive orders to speed up vaccine distribution ...  \\ 
      \hline
    \end{tabular}     
  }
  \caption{
  Two sample CNN news articles in 2015 and 2021, respectively, as well as their summaries generated by the PEGASUS~\cite{DBLP:conf/icml/ZhangZSL20} model that is fine-tuned on the CNN/DM dataset. 
  We highlight the entity mentions of Joe Biden in the articles and summary in red color. We underline the factual error in the summaries. 
  }
    \label{fig:intro-example}
\end{figure}

In Figure~\ref{fig:intro-example}, we present two CNN news articles in 2015 and 2021, respectively, as well as their summaries generated by the PEGASUS~\cite{DBLP:conf/icml/ZhangZSL20} model that is fine-tuned on the CNN/DM dataset. 
From the summary of the article in 2015, we observe that PEGASUS generates ``Vice President Joe Biden'', which is faithful to the source document. 
However, since both the training set of CNN/DM and the pre-training corpora of PEGASUS contain text data in 2015, it is not clear whether the model utilizes the document content or the \textit{parametric world knowledge that is memorized during pre-training and fine-tuning} to generate the correct information about Joe Biden.

On the other hand, the news article in 2021 from Figure~\ref{fig:intro-example} indicates that Joe Biden is a president. 
Since both the CNN/DM training set and standard pre-training corpora do not contain data after 2019, the PEGASUS model must leverage the document content to generate the expression ``President Joe Biden''. 
Unfortunately, the PEGASUS model hallucinates the content of ``Vice President Joe Biden'' in the summary which is unfaithful. 

Based on the above qualitative results, we suspect that the promising performance of PLMs in existing summarization benchmarks relies on their parametric world knowledge. 
In order to reliably evaluate the temporal generalization ability of pre-trained summarization models, the source articles in an evaluation benchmark should contain \textit{knowledge conflicts}, i.e., the articles should present knowledge that contradicts the parametric world knowledge in PLMs. 

To this end, we propose the \textsc{TempoSum} benchmark, which contains text articles with \textbf{knowledge conflicts} to assess the temporal generalization capability of text summarization models. 
Specifically, our benchmark consists of two summarization datasets of different abstractiveness collected from CNN and BBC News ranging from 2010 to 2022. 
For each dataset, we construct a future test set, in which all the articles are published after 2019 and contain knowledge conflicts to evaluate the robustness of pre-trained summarization on articles from a future temporal period. 

Based on our benchmark, we present the first systematic human evaluation of the temporal generalization ability of text summarization models.
This paper aims to answer two important research questions. 
First, \textbf{how do the pre-training process and fine-tuning datasets affect the temporal generalization abilities of summarization models?}
We compare the performance of the SOTA pre-trained text summarization model, PEGASUS, and a non-pretrained Transformer model on our proposed benchmark. 
Our results demonstrate that although the pre-training process of PEGASUS helps the model generates more meaningful and grammatical summaries, it also encourages the model to hallucinate outdated information according to its parametric world knowledge. 
Moreover, we observe that the divergence between the source documents and their reference summaries in the finetuning dataset causes summarization models to rely on their parametric world knowledge instead of the source text. 

Second, \textbf{are recent faithfulness enhancement methods and faithfulness evaluation methods effective on future data?}
We first analyze the performance of two recent faithfulness enhancement methods, CLIFF~\cite{DBLP:conf/emnlp/Cao021} and ENT~\cite{DBLP:journals/corr/abs-2209-03479}, which are built on contrastive learning and copy mechanism, respectively. 
We find that the contrastive learning-based method outperforms the copy-based method in terms of faithfulness improvement on future data. 
However, the performance of these two faithfulness enhancement methods heavily depends on the domain of the data. 
Namely, both of these methods fail to improve the faithfulness of the text summarization in future CNN news articles. 
Then, we investigate the human correlations of two popular faithfulness evaluation metrics, FactCC~\cite{DBLP:conf/emnlp/KryscinskiMXS20} and QAFactEval~\cite{DBLP:conf/naacl/FabbriWLX22}, on future data. The results show that both of these metrics have a weak correlation with human judgments.

Taken together, it is essential to use temporally split test sets that contain knowledge conflicts to accurately estimate the temporal generalization performance of text summarization models. 
Our findings suggest that PLMs are prone to rely on their parametric knowledge and exhibit a tendency to generate hallucinations containing outdated knowledge. Therefore, a new evaluation protocol is required to estimate the extent to which PLMs rely on their parametric world knowledge.

\section{Related Work}

\subsection{Text Summarization Datasets}

Most of the existing text summarization datasets, such as CNN/Dailymail~\cite{DBLP:conf/nips/HermannKGEKSB15, DBLP:conf/conll/NallapatiZSGX16}, XSum~\cite{DBLP:conf/emnlp/NarayanCL18}, Newsroom~\cite{DBLP:conf/naacl/GruskyNA18}, focus on single document summarization in the news domain.
The Multi-News dataset~\cite{DBLP:conf/acl/FabbriLSLR19} is proposed for multi-document news summarization, in which the input is a set of news documents. 
The SumREN dataset~\cite{DBLP:journals/corr/sumren22} is constructed for reported speech summarization that aims to summarize the reported statements made by a specific person in news documents. 
Other text summarization datasets focused on summarizing scientific articles~\cite{DBLP:conf/naacl/CohanDKBKCG18,DBLP:conf/emnlp/LuDC20}, legal documents~\cite{kornilova-eidelman-2019-billsum}, events~\cite{DBLP:conf/acl/wcep20,DBLP:conf/www/pdsum23,li2021timeline}, or dialogues~\cite{gliwa-etal-2019-samsum,DBLP:conf/naacl/ZhuLMZ21,DBLP:journals/kbs/ChenLCK21}. 
However, the test sets of the above datasets do not evaluate the temporal generalization capability of summarization models. A realistic evaluation benchmark should include data from different temporal periods to ensure an accurate evaluation. As most of the existing PLMs are pre-trained on corpora before 2019, \textsc{TempoSum} includes the data collected from 2010 to 2022. The extended collection periods allow the research community to explore the temporal generalization ability of these models. 
Recently, \cite{DBLP:journals/corr/abs-2209-12356} collect 100 news articles in 2022 from CNN and BBC, but they do not select articles that contain knowledge conflicts. In contrast, our benchmark comprises recent news articles that contain knowledge conflicts with existing PLMs, thereby enabling us to examine the temporal generalization ability of PLMs. 

\subsection{Evaluation of Faithfulness and Factuality} 
There are several datasets that annotate the faithfulness of generated summaries. 
Some of these datasets~\cite{DBLP:conf/emnlp/KryscinskiMXS20, DBLP:conf/acl/WangCL20,DBLP:journals/tacl/FabbriKMXSR21} annotate an overall faithfulness label for each summary. \citet{pagnoni-etal-2021-understanding, DBLP:conf/naacl/AnnotatingFineGrained21,DBLP:conf/emnlp/HuangCYBWXZ20} define a more fine-grained typology of faithfulness errors, and different models~\cite{DBLP:conf/acl/FineGrainFact23} are proposed to detect the factual error types. 
Other datasets~\cite{DBLP:conf/acl/MaynezNBM20,DBLP:conf/emnlp/Cao021,DBLP:conf/acl/CaoDC22} annotate factuality labels to indicate whether the generated summaries are factually correct. 
All the above datasets annotate summaries in the existing CNN/DM and XSum benchmarks. By contrast, we collect recent news articles that contain knowledge conflicts to study the temporal generalization ability of PLMs.

\subsection{Out-of-distribution Generalization of Language Models}
Some previous works~\cite{DBLP:journals/jmlr/RaffelSRLNMZLL20,DBLP:journals/corr/abs-2211-03154,DBLP:conf/acl/GrangierI22} study the generalization capability of language models (LMs) in out-of-domain data. 
Other studies~\cite{lazaridou2021mind, DBLP:conf/naacl/LuuKGMS22, DBLP:conf/emnlp/RottgerP21, DBLP:conf/wsdm/RosinGR22, DBLP:conf/acl/LoureiroBNAC22, DBLP:journals/tacl/AgarwalN22} analyze the temporal generalization ability of LMs on future test sets that do not have temporal overlap with the training set and standard pre-training corpora, but these studies do not analyze the effects of parametric world knowledge on the temporal generalization ability of PLMs. 
Recently, \citet{DBLP:conf/emnlp/LongprePCRD021} propose a method to artificially generate data samples that contain knowledge conflicts and study how parametric world knowledge affects the performance of question-answering models. 
In contrast, our method selects recent news articles that present knowledge conflicts. Thus, the distribution of the samples created by our method is more similar to the distribution of real data. 
Moreover, we conduct extensive human evaluations to analyze the types of errors made by pre-trained summarization models.

\section{\textsc{TempoSum} Benchmark}

We first propose the \textsc{TempoSum} benchmark to investigate the temporal generalization capability of PLMs on future documents that present knowledge conflicts, i.e., the scenarios where the knowledge presented in the source document contradicts the parametric world knowledge in PLMs. 

\subsection{Data Source}
\label{sess:data-sources}

Previous pre-trained summarization models are mostly fine-tuned on the XSum~\cite{DBLP:conf/emnlp/NarayanCL18} or CNN/DM~\cite{DBLP:conf/nips/HermannKGEKSB15} datasets. 
To evaluate the temporal generalization ability of PLMs that are finetuned on XSum and CNN/DM, we create the following two datasets that contain articles from an extended time period. 
We filter out all the samples that overlap with the training sets of XSum and CNN/DM.

\paragraph{BBC dataset:} 
The existing XSum dataset consists of news articles from BBC news between 2010 to 2017. The reference summaries in XSum are highly abstractive. 
In our benchmark, we crawl news articles from BBC news between 2010 to 2022 to construct the BBC dataset.

\paragraph{CNN dataset:} 
The existing CNN/DM dataset contains news articles from CNN news and Dailymail news between 2007 and 2015. The reference summaries in CNN/DM are highly extractive. 
We collect news articles from CNN news\footnote{We exclude DailyMail from our benchmark because this news agency is reported to have low credibility~\cite{DBLP:journals/corr/abs-2209-12356} 
} 
between 2012 to 2022 to construct the CNN dataset in our benchmark. 

\subsection{Test Set Construction}
\label{sess:identify-conflicted-knowledge}

To understand the temporal generalization ability of pre-trained summarization models, 
we construct a \textbf{future test set} and an \textbf{in-distribution test set} for each dataset in our benchmark. 
A future test set consists of samples that present knowledge conflicts, while an in-distribution test set comprises samples that are consistent with the parametric knowledge of pre-trained summarization models. 
To construct the test sets in our benchmark, we first collect \textbf{evolving facts} that change over time. 
Next, we assume that the collected evolving facts that occurred after 2019 are \textbf{knowledge conflicts} and use them to select news articles. 

\subsubsection{Identifying evolving facts}
We use Wikidata \cite{DBLP:journals/cacm/VrandecicK14} as the knowledge base to collect evolving facts. Wikidata store facts in the form of (\texttt{subject}, \texttt{relation}, \texttt{object}) triples. 
Since standard pre-training corpora only obtain data up to 2019, we assume that any fact that is changed after 2019 is unaware by PLMs. 
Concretely, if the object of a (\texttt{subject}, \texttt{relation}) pair is changed after 2019, then all the facts that contain the (\texttt{subject}, \texttt{relation}) pair are considered as evolving facts. 
For example, Wikidata contains a fact (\texttt{Joe Biden}, \texttt{position-held}, \texttt{Vice President of the United States}) since 2009. 
However, the object of (\texttt{Joe Biden}, \texttt{position-held}) is changed to ``\texttt{President of the United States}'' in 2021. Thus, both the facts 
(\texttt{Joe Biden}, \texttt{position-held}, \texttt{Vice President of the United States}) and (\texttt{Joe Biden}, \texttt{position-held}, \texttt{President of the United States}) are considered as evolving facts.  

As the space of evolving facts is extremely large, our work focuses on the facts that are related to politicians as a proxy to study the effects of conflicted knowledge on PLMs.
Specifically, we only extract evolving facts that have the ``\texttt{position-held}'' or ``\texttt{position-held-by}'' relation. 
We select these two relations because the standard text summarization fine-tuning datasets (e.g., CNN/DM, XSum) focus on the news domain, and most of their samples contain politicians.
Moreover, as the positions of government officials often change, the facts that contain these two relations have \textit{strong temporal dynamics} \cite{lazaridou2021mind, DBLP:journals/tacl/DhingraCEGEC22}. 

\subsubsection{Selecting articles with evolving facts}
For both BBC and CNN datasets in our benchmark, we only crawl the news articles in which the salient content contains the subject or object entity of an evolving fact\footnote{We crawl archived BBC and CNN articles from Internet Archive (\href{https://web.archive.org/}{https://web.archive.org/})}. 
It is because PLMs prone to rely on their parametric knowledge when summarizing the entities that are related to evolving facts, as shown in Figure~\ref{fig:intro-example}. 
Specifically, all the crawled articles satisfy the following two criteria: (1) the article includes both the subject and object of at least one evolving fact; (2) the reference summary includes either the subject or the object of an evolving fact. In this way, all our crawled articles that are published after 2019 should contain facts that contradict the knowledge stored in PLMs (\textbf{knowledge conflicts}). 
For each of the BBC and CNN datasets, we partition the crawled articles into {future test set} and {in-distribution test set} according to their publication time. The crawled articles that were published after 2019 belong to the future test set; the remaining articles belong to the in-distribution test set. 
We present the statistics of our constructed test sets in Table \ref{tab:dataset-stat}.

\subsection{Human Evaluation Protocol}

One major harmful effect of parametric world knowledge is that it may cause summarization models to hallucinate contents that are not faithful to the source document, as shown in Figure~\ref{fig:intro-example}. 
To understand how parametric world knowledge affects the faithfulness of the generated summaries, we define three fine-grained categories of hallucinations, which include  \textbf{outdated hallucinations}, \textbf{non-verifiable hallucinations} and \textbf{factual hallucination}. Table~\ref{tab:factual-error-examples} presents the complete definition of our proposed typology of hallucinations. In contrast, previous works~\cite{DBLP:conf/acl/MaynezNBM20} only determine whether the hallucinations are factual or non-factual. 

For each generated summary, we first decide whether the sentence is faithful, i.e., the information can be entailed by the source article. If it is not faithful, we verify it through Wikipedia to decide whether they are factual to world knowledge at the time the article was published or outdated if it could be connected to the knowledge from the past. If there is no information found to support the unfaithful entities, then labeled as non-verifiable errors. The complete human evaluation instruction is in Appendix~\ref{sec:appendix-human-evaluation}.

\begin{table}[t]
 \setlength{\tabcolsep}{9pt}
\centering \small
\begin{tabular}{cccr}
\toprule
\textbf{Dataset}  & \textbf{Test set} & \textbf{Year} & \textbf{\# Samples} \\ 
\midrule
 \multirow{2}{*}{BBC} & In-dist. & 2010-2017 & 6,254  \\
\cmidrule{2-4}
  & Future & 2020-2022  & 2,260 \\
 \midrule
 \multirow{2}{*}{CNN} & In-dist. & 2012-2015  & 970 \\
\cmidrule{2-4}
  & Future & 2020-2022  & 3,250  \\
\bottomrule
\end{tabular}
\caption{\label{tab:dataset-stat} 
The dataset statistics of \textsc{TempoSum}. 
Our evaluation benchmark consists of two datasets: BBC and CNN.
We construct an in-distribution (In-dist.) and a future test set for each dataset.
}
\end{table}

\begin{table*}[ht]
    \small
    \begin{tabular}{  p{2.9cm} | p{6cm} | p {5.6cm} }
        \toprule
\textbf{Category} & \textbf{Definition} & \textbf{Example} \\
\midrule
  Outdated hallucinations & Hallucinations that are consistent with world knowledge before the time in which the article is published.  & \textcolor{red}{Vice President} Joe Biden has signed executive orders to speed up vaccine distribution. \\
\midrule
 Non-verifiable hallucinations & Hallucinations that cannot be verified by world knowledge at any time stamp. If a summary contains totally ungrammatical sentences, we will also consider the grammatical errors as non-verifiable hallucinations. 
 &  President \textcolor{red}{Thomas Biden} has signed executive orders to speed up vaccine distribution. \\
 \midrule
 Factual hallucinations & Hallucinations that are consistent with world knowledge at the time the article is being published. &  \textcolor{red}{US} President Joe Biden has signed executive orders to speed up vaccine distribution. 
 \\ 
\midrule
    \end{tabular}
    \caption{Our proposed typology for hallucinations in system-generated summaries. 
    The word spans that are inconsistent with the source text (hallucinations) are highlighted in \textcolor{red}{red} color. 
    The truncated source article (published in 2021) for the examples: 
    \textit{President Joe Biden’s mission to speed up the lackluster rollout of coronavirus vaccines. Biden signed a slew of executive orders this week aimed at accelerating the distribution of vaccine and tests.
    }
    }
    \label{tab:factual-error-examples}
\end{table*}

\section{Experiment Settings}

\subsection{Summarization Models}
We perform extensive experiments using the following models. 
More implementation details are described in Appendix \ref{sess:implementation-details}.

\paragraph{PEGASUS:} The state-of-the-art pre-trained text summarization model. This model is a Transformer model that is pre-trained using the gap-sentence prediction objective~\cite{DBLP:conf/icml/ZhangZSL20}. 

\paragraph{Transformer:} 
To study how the pre-training process affects the temporal generalization performance of summarization models, 
we also evaluate the performance of a non-pretrained vanilla Transformer model~\cite{vaswani2017attention} in our experiments. 

\paragraph{CLIFF:} A faithfulness enhancement method that applies contrastive learning to teach a summarization model to distinguish faithful summaries from unfaithful summaries~\cite{DBLP:conf/acl/CaoDC22}. This method implicitly encourages text summarization models to generate faithful summaries. We apply the CLIFF method to the PEGASUS model in our experiments. 

\paragraph{ENT:} A faithfulness enhancement method that encourages summarization models to explicitly copy salient entities from the source text via an entity span copy mechanism \cite{DBLP:journals/corr/abs-2209-03479}. We apply the ENT method to the PEGASUS model in our experiments.

\subsection{Human Evaluation}

We conduct extensive human evaluations to annotate the types of hallucinations made by system-generated summaries\footnote{We use the annotation tool proposed by \citet{doccano} to conduct our human evaluation.}. 
Four authors from this paper annotate the above model-generated summaries. 
We annotate 150 random articles from each test split (a total of 600 articles for both the BBC and CNN datasets). Each sample is annotated by two of our authors. 
The matching rate of annotations on the in-distribution test set was 68\% and 76\% for the BBC and CNN datasets. As for the future test set, the annotations matched 72\% and 85\% for the BBC and CNN datasets, respectively.

\subsection{Automatic Evaluation Metrics}
\paragraph{FactCC:}  
An entailment-based faithfulness evaluation metric proposed by \citet{DBLP:conf/emnlp/KryscinskiMXS20}. Previous studies~\cite{pagnoni-etal-2021-understanding} show that FactCC has a strong correlation with human judgments on existing benchmarks. 

\paragraph{QAFactEval:} The state-of-the-art faithfulness evaluation for text summarization~\cite{DBLP:conf/naacl/FabbriWLX22}. This metric utilizes question-generation and question-answering models to estimate the faithfulness of a summary. 

\section{Result Analysis}

\subsection{How do the pre-training process and fine-tuning datasets affect the temporal generalization abilities of summarization models?}

\begin{figure}[th]
    \centering
    \begin{subfigure}{\columnwidth}
    \centering
    \includegraphics[width=\linewidth]{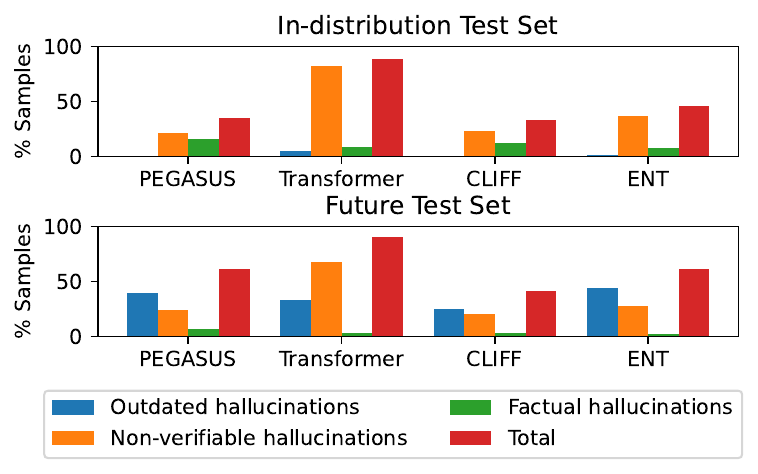} 
    \caption{BBC Dataset}
    \label{fig:human-anntations-bbc}
    \end{subfigure}
    
    \begin{subfigure}{\columnwidth}
    \centering
    \includegraphics[width=\linewidth]{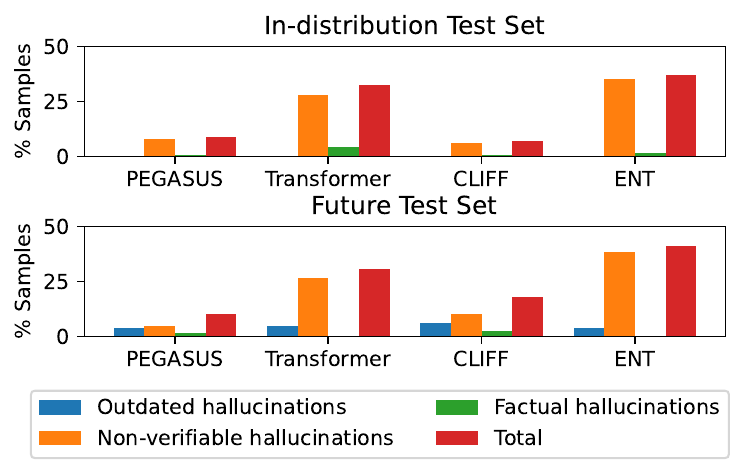} 
    \caption{CNN Dataset}
    \label{fig:human-anntations-cnn}
    \end{subfigure}

    \caption{Human evaluation results on the types of hallucinations made by different models on the \textsc{TempoSum} benchmark. 
    }
    \label{fig:human-annotations}
\end{figure}

We illustrate the distribution of hallucinations made by different models on the BBC and CNN datasets in Figure~\ref{fig:human-anntations-bbc} and Figure~\ref{fig:human-anntations-cnn}, respectively. From these two figures, we draw the following observations. 

\paragraph{Pre-training improves the faithfulness of a summarization model on future data. }
We observe that PEGASUS produces significantly fewer hallucinations than its non-pretrained counterpart in the future test sets of both BBC and CNN datasets. The results suggest that pre-training helps a summarization model generates more faithful summaries even if the source articles contain knowledge conflicts.

\paragraph{Pre-training encourages a summarization model to hallucinate outdated information based on its parametric knowledge. }

We observe that on the future test set of BBC, PEGASUS generates substantially more outdated hallucinations than Transformer (39.81\% vs. 33.33\%). While on the future test set of CNN, PEGASUS and Transformer obtain a similar number of outdated hallucinations (3.94\% vs 4.72\%). 
Our human evaluation results reveal that the parametric world knowledge learned during pre-training harms the faithfulness performance of PEGASUS on future test sets. 
Previous studies mainly consider the benefits of parametric knowledge \cite{DBLP:conf/acl/CaoDC22, DBLP:conf/acl/MaynezNBM20}, but few of them critically study its negative impacts.
Our results indicate that parametric knowledge can be harmful and may guide the model to generate unfaithful information.

\paragraph{The source-target divergence in the finetuning dataset encourages the model to generate summaries with less reliance on the source.} 

\begin{figure}[t!]
    \footnotesize
    \begin{tabular}{p{1.4cm} p{5.6cm}} 
    \toprule
    \textbf{Source article} &
    The \textcolor{red}{prime minister} was speaking in the Commons two days after announcing the lockdown at a televised press conference...
    UK recorded 18,950 new confirmed cases of coronavirus and 136 deaths ...
    \textcolor{red}{Boris Johnson} is stuck in between two groups... \\
    \midrule
    \textbf{Summary by PEGASUS} &
    \textcolor{red}{Prime Minister} \textcolor{red}{\underline{David Cameron}} has defended the government 's response to the outbreak of the coronavirus, which has killed 136 people in the UK. \\
    \midrule
    \textbf{Summary by CLIFF} &
    \textcolor{red}{Prime Minister} \textcolor{red}{Boris Johnson} has defended the government's response to the outbreak of coronavirus in the UK. \\
    \bottomrule
    \end{tabular}
    \vspace{-0.2cm}
   \caption{An example of a BBC news article from 2020, and the summaries generated by PEGASUS and CLIFF. 
   We highlight the entity mentions of Prime Minister and Boris Johnson in the article and the generated summaries in \textcolor{red}{red} color. We underline the factual error in the summaries.
   }
    \label{fig:bbc-samples}
    \vspace{-0.1cm}
\end{figure}

Our evaluation results show that models that are fine-tuned on XSum generate more hallucinations on future data.
The reason is that a significant portion of reference summaries in the XSum dataset have the entity missing problem~\cite{tejaswin-etal-2021-well} (i.e., reference summaries contain entities that do not appear in the source text). 
Thus, models trained on this dataset are encouraged to hallucinate contents that are not inferable from the source article. 

We observe that models that are fine-tuned on the XSum dataset learn \textbf{spurious correlations} from the training data. Namely, these models learn to generate entities that frequently co-occur on the training data, e.g., Prime Minister and David Cameron.  
Figure \ref{fig:bbc-samples} shows an example of a BBC news article published in 2020. The content of the article implies that Boris Johnson is the prime minister of the UK. 
However, PEGASUS generates ``Prime Minister David Cameron'' in the summary, which is an outdated hallucination. This example suggests that PEGASUS memorizes a strong correlation between Prime Minister and David Cameron from the finetuning dataset. 
Learning such a spurious correlation is undesirable since a text summarization model should learn to ground the generation decision on the facts presented in the source text. 
Moreover, this kind of behavior harms the temporal generalization ability of text summarization since the knowledge of our world evolves constantly. 

\paragraph{The pre-training process helps generate meaningful and grammatical summaries.}
On the BBC dataset, PEGASUS generates much fewer non-verifiable errors than Transformer for both the in-distribution test set (21.36\% vs. 82.52\%) and future test set (24.07\% vs. 67.59\%). 
Similarly, on the CNN dataset, PEGASUS produces less non-variable error than Transformer for both the in-distribution test set (7.89\% vs. 28.07\%) and future test set (4.72\% vs. 26.77\%). 
The reason may lie in that the pre-training process of PEGASUS allows the model to learn more general linguistic features and time-invariant world knowledge (e.g., London is the capital city of the United Kingdom), which can be generalized to the data from different temporal periods.

\paragraph{Summary.}
Our experiment results show that fine-tuning datasets plays a crucial role in the temporal generalization performance of text summarization models.
The divergence between the source document and reference summary
encourages models to heavily rely on their parametric knowledge, limiting their temporal generalization ability. 
On the other hand, outdated hallucination is a critical concern for the temporal generalization ability. It becomes a non-neglectable factor in the faithfulness performance on the future test set.

\subsection{Are recent faithfulness enhancement methods and faithfulness evaluation methods effective on future data?}

We first investigate the temporal generalization of two recent faithfulness enhancement methods, CLIFF and ENT. 
Then, we evaluate the human correlations of two popular faithfulness metrics, FactCC and QAFactEval.

\subsubsection{Performance of the CLIFF method}

Figure \ref{fig:human-annotations} shows the annotation results of CLIFF on both datasets. 
We have the following observations. 

\paragraph{CLIFF effectively prevents summarization models from relying on their parametric knowledge on the BBC dataset. }

We observe that CLIFF generates significantly fewer hallucinations than PEGASUS on the future test set of BBC (41.67\% vs. 61.11\%). 
The remarkable faithfulness improvement of CLIFF is attributed to its effectiveness in encouraging models not to rely on their parametric knowledge. 
On the future test set, CLIFF generated fewer summaries containing outdated hallucinations than PEGASUS (25.00\% vs 39.81\%). 
Moreover, CLIFF shows its robustness in information fusion, Figure \ref{fig:bbc-samples} shows an example of a BBC article from 2020 and the summary generated by CLIFF.
It successfully fuses the information from the source article and generates a faithful summary with new knowledge ``Prime Minister Boris Johnson'', while PEGASUS utilizes its parametric knowledge to generate a summary with outdated hallucination.

\paragraph{The effectiveness of CLIFF on future data highly depends on the finetuning dataset.}
Different from the competitive performance observed on the BBC dataset, CLIFF has negative impacts on the CNN dataset.
CLIFF slightly improves the faithfulness performance on the in-distribution test set, producing fewer hallucinations than PEGASUS (6.14\% vs. 7.89\%). 
CLIFF is ineffective on the future test set and generates more hallucinations than PEGASUS (18.11\% vs. 10.24\%).
Moreover, the training objective of CLIFF fails to mitigate the reliance on parametric knowledge and yields more outdated hallucinations than PEGASUS (6.30\% vs. 3.94\%).

While CLIFF performs well on the in-distribution test set, its performance on the future test set is inconsistent across the two datasets in the \textsc{TempoSum} benchmark. 
It remains challenging to develop faithfulness enhancement methods that are generalizable to the future.

\subsubsection{Performance of the ENT method}

Figure \ref{fig:human-annotations} shows the annotation results for the ENT method on both datasets. We draw the following observations. 

\paragraph{The ENT method fails to improve the faithfulness of summarization models on future data.}

From Figure \ref{fig:human-annotations}, we observe that ENT and PEGASUS produce a similar amount of hallucinated content (61.11\% vs. 61.11\%) on the future test set of the BBC dataset. Whereas in the future test set of CNN, the ENT model produces more hallucinations than the PEGASUS model (40.94\% vs. 10.24\%). 
Moreover, ENT does not reduce the tendency of a model to rely on its parametric world knowledge. 
In comparison to PEGASUS, ENT produces a higher percentage of outdated hallucinations on the future test set of BBC (44.44\% vs. 39.81\%). 
Whereas in the future test set of the CNN dataset, ENT produces the same level of outdated hallucinations as PEGASUS (3.93\% vs. 3.93\%). 

We suspect that the ineffectiveness of ENT is due to the restrictions in its copy mechanism. The copy mechanism of ENT only copies named entities from the source text that are identified by a NER tool. However, unnamed entities like government officials (e.g., the prime minister) often occur in the source text and reference summaries in our benchmark. This type of unnamed entity cannot be recognized by NER and is excluded from the copy mechanism of ENT.

\begin{table}[t]
\centering \small
\begin{tabular}{lccc}
\toprule
\textbf{Source}  &  \textbf{Metric} & \textbf{In-distribution} & \textbf{Future}  \\ 
\midrule
 \multirow{2}{*}{BBC} & FactCC & -0.03 & 0.138 \\
 & QAFactEval & \textbf{0.406} & \textbf{0.317} \\
 \midrule
 \multirow{2}{*}{CNN} &  FactCC & 0.219 & 0.235 \\
 & QAFactEval & \textbf{0.303} & \textbf{0.377} \\
\bottomrule
\end{tabular}
\caption{\label{tab:corelation-human-auto} 
Spearman's rank correlation between human judgments and faithfulness metrics in different test sets. 
}
\end{table}

\subsubsection{Human correlations results of faithfulness evaluation metrics}
\label{sess:automatic-evaluation}

\paragraph{Recent faithfulness evaluation metrics demonstrate weak correlations with human judgments on future data. }
We present the human correlation results of FactCC and QAFactEval on the \textsc{TempoSum} benchmark in Table~\ref{tab:corelation-human-auto}. 
Although QAFactEval consistently obtains higher human correlations than FactCC on the future test sets of CNN and BBC, the Spearman's correlation coefficient scores \cite{myers2004spearman} obtain by these two metrics are lower than 0.4, suggesting that these metrics only have weak correlations with human judgments. Thus, we still need to develop more accurate faithfulness evaluation metrics for future data. 

\subsubsection{Summary}
Enhancing and evaluating the faithfulness of summarization models to future data poses a significant challenge. 
Current faithfulness enhancement and evaluation methods are typically developed based on observations made from existing evaluation benchmarks, which may not be generalizable to future data. 
Instead, we argue that the use of non-temporal overlap samples for evaluation purposes can more accurately reflect the generalization capability of summarization systems and metrics, while also providing a fair comparison of performance between different faithfulness enhancement and evaluation methods.

\section{Recommendations}

Based on the observations from our experiments, we recommend the following four takeaways to the summarization research community.

\begin{enumerate}[nolistsep,wide]
\item We should use temporally split test sets that contain knowledge conflicts to evaluate the performance of summarization systems. 
This kind of test set allows us to isolate the effects of parametric world knowledge to accurately evaluate the temporal generalization performance of text summarization.  
\item Human evaluations for pre-trained summarization models should annotate the number of generated summaries that contain outdated hallucinations. This practice allows us to assess the extent to which pre-trained summarization models rely on their parametric world knowledge. 
\item In order to improve the temporal generalization capability of summarization models, we need to develop techniques to prevent summarization models from learning spurious correlations among entities during pre-training and fine-tuning. Our experiments show that spurious correlation is a key factor that affects the temporal generalization ability of summarization models on future data. 
\item Users need to be cautious when using existing automatic metrics to evaluate the faithfulness of summaries on future data. The community should develop accurate automatic faithfulness metrics that can generalize well on future data. 
\end{enumerate}

\section{Conclusion}

In this paper, we propose \textsc{TempoSum}, a novel evaluation benchmark for assessing the temporal generalization capability of text summarization models.  
We collect samples from a broad temporal range which allow us to study the performance of PLMs on data from different temporal periods. 
Through extensive experiments on \textsc{TempoSum}, we find that PLMs are vulnerable to generating summaries with outdated hallucinations and ignore the supporting evidence provided by source articles.
It is crucial to develop techniques to prevent summarization systems from generating outdated information and design automatic evaluation metrics to detect outdated hallucinations. 
\textsc{TempoSum} provides a valuable data corpus which covers a wide range of temporal periods, along with human judgements from various summarization systems, to support future studies in this direction.

\section*{Limitations}
Our study of temporal generalization is limited to news domain datasets. 
In the future, we still need empirical studies for the temporal generalization abilities of text summarization models on other domains (e.g., dialog and scientific domains). 
Moreover, our analysis of knowledge conflicts mainly focuses on entity relationships with politicians. We can consider other categories of entities in future work. 
Furthermore, as most of the existing PLMs (e.g., PEGASUS, BART, T5, etc.) are pre-trained on corpora before 2019, \textsc{TempoSum} is designed to evaluate the temporal generalization ability of PLMs that were pre-trained on corpora before 2019. To access the temporal generalization capability of PLMs trained on more recent data (e.g., data from 2019 to 2021), we need to construct an evaluation benchmark that includes data from future periods (e.g, data after 2021).

\section*{Ethics Statement}
This work proposes a new benchmark for accessing the temporal generalization ability of text summarization models. 
We only collect news articles from creditableg news agencies to avoid including offensive content in our benchmark.  
The news articles used in our benchmark are publicly available on the internet and therefore our benchmark does not reveal any private information about individuals.

\section*{Acknowledgements}

We would like to thank the anonymous reviewers and meta-reviewer for their valuable feedback on this work.
This work was supported in part by the Science and Technology Development Fund, Macau SAR (Grant Nos. FDCT/060/2022/AFJ, FDCT/0070/2022/AMJ) and the Multi-year Research Grant from the University of Macau (Grant No. MYRG-GRG2023-00006-FST-UMDF). Xuebo Liu was supported by the CCF-Tencent Rhino-Bird Open Research Fund. This work was performed in part at SICC which is supported by SKL-IOTSC, and HPCC supported by ICTO of the University of Macau. 


\bibliography{anthology,custom}

\begin{thebibliography}{47}
\expandafter\ifx\csname natexlab\endcsname\relax\def\natexlab#1{#1}\fi

\bibitem[{Agarwal and Nenkova(2022)}]{DBLP:journals/tacl/AgarwalN22}
Oshin Agarwal and Ani Nenkova. 2022.
\newblock \href {https://transacl.org/ojs/index.php/tacl/article/view/3863} {Temporal effects on pre-trained models for language processing tasks}.
\newblock \emph{Trans. Assoc. Comput. Linguistics}, 10:904--921.

\bibitem[{Cao et~al.(2022)Cao, Dong, and Cheung}]{DBLP:conf/acl/CaoDC22}
Meng Cao, Yue Dong, and Jackie Chi~Kit Cheung. 2022.
\newblock \href {https://doi.org/10.18653/v1/2022.acl-long.236} {Hallucinated but factual! inspecting the factuality of hallucinations in abstractive summarization}.
\newblock In \emph{Proceedings of the 60th Annual Meeting of the Association for Computational Linguistics (Volume 1: Long Papers), {ACL} 2022, Dublin, Ireland, May 22-27, 2022}, pages 3340--3354. Association for Computational Linguistics.

\bibitem[{Cao and Wang(2021)}]{DBLP:conf/emnlp/Cao021}
Shuyang Cao and Lu~Wang. 2021.
\newblock \href {https://doi.org/10.18653/v1/2021.emnlp-main.532} {{CLIFF:} contrastive learning for improving faithfulness and factuality in abstractive summarization}.
\newblock In \emph{Proceedings of the 2021 Conference on Empirical Methods in Natural Language Processing, {EMNLP} 2021, Virtual Event / Punta Cana, Dominican Republic, 7-11 November, 2021}, pages 6633--6649. Association for Computational Linguistics.

\bibitem[{Chan et~al.(2023)Chan, Zeng, and Ji}]{DBLP:conf/acl/FineGrainFact23}
Hou~Pong Chan, Qi~Zeng, and Heng Ji. 2023.
\newblock \href {https://doi.org/10.18653/v1/2023.findings-acl.402} {Interpretable automatic fine-grained inconsistency detection in text summarization}.
\newblock In \emph{Findings of the Association for Computational Linguistics: {ACL} 2023, Toronto, Canada, July 9-14, 2023}, pages 6433--6444. Association for Computational Linguistics.

\bibitem[{Chen et~al.(2021)Chen, Li, Chan, and King}]{DBLP:journals/kbs/ChenLCK21}
Wang Chen, Piji Li, Hou~Pong Chan, and Irwin King. 2021.
\newblock \href {https://doi.org/10.1016/j.knosys.2021.107328} {Dialogue summarization with supporting utterance flow modelling and fact regularization}.
\newblock \emph{Knowl. Based Syst.}, 229:107328.

\bibitem[{Cohan et~al.(2018)Cohan, Dernoncourt, Kim, Bui, Kim, Chang, and Goharian}]{DBLP:conf/naacl/CohanDKBKCG18}
Arman Cohan, Franck Dernoncourt, Doo~Soon Kim, Trung Bui, Seokhwan Kim, Walter Chang, and Nazli Goharian. 2018.
\newblock \href {https://doi.org/10.18653/v1/n18-2097} {A discourse-aware attention model for abstractive summarization of long documents}.
\newblock In \emph{Proceedings of the 2018 Conference of the North American Chapter of the Association for Computational Linguistics: Human Language Technologies, NAACL-HLT, New Orleans, Louisiana, USA, June 1-6, 2018, Volume 2 (Short Papers)}, pages 615--621. Association for Computational Linguistics.

\bibitem[{Dhingra et~al.(2022)Dhingra, Cole, Eisenschlos, Gillick, Eisenstein, and Cohen}]{DBLP:journals/tacl/DhingraCEGEC22}
Bhuwan Dhingra, Jeremy~R. Cole, Julian~Martin Eisenschlos, Daniel Gillick, Jacob Eisenstein, and William~W. Cohen. 2022.
\newblock \href {https://doi.org/10.1162/tacl\_a\_00459} {Time-aware language models as temporal knowledge bases}.
\newblock \emph{Trans. Assoc. Comput. Linguistics}, 10:257--273.

\bibitem[{Fabbri et~al.(2021)Fabbri, Kryscinski, McCann, Xiong, Socher, and Radev}]{DBLP:journals/tacl/FabbriKMXSR21}
Alexander~R. Fabbri, Wojciech Kryscinski, Bryan McCann, Caiming Xiong, Richard Socher, and Dragomir~R. Radev. 2021.
\newblock \href {https://doi.org/10.1162/tacl\_a\_00373} {Summeval: Re-evaluating summarization evaluation}.
\newblock \emph{Trans. Assoc. Comput. Linguistics}, 9:391--409.

\bibitem[{Fabbri et~al.(2019)Fabbri, Li, She, Li, and Radev}]{DBLP:conf/acl/FabbriLSLR19}
Alexander~R. Fabbri, Irene Li, Tianwei She, Suyi Li, and Dragomir~R. Radev. 2019.
\newblock \href {https://doi.org/10.18653/v1/p19-1102} {Multi-news: {A} large-scale multi-document summarization dataset and abstractive hierarchical model}.
\newblock In \emph{Proceedings of the 57th Conference of the Association for Computational Linguistics, {ACL} 2019, Florence, Italy, July 28- August 2, 2019, Volume 1: Long Papers}, pages 1074--1084. Association for Computational Linguistics.

\bibitem[{Fabbri et~al.(2022)Fabbri, Wu, Liu, and Xiong}]{DBLP:conf/naacl/FabbriWLX22}
Alexander~R. Fabbri, Chien{-}Sheng Wu, Wenhao Liu, and Caiming Xiong. 2022.
\newblock \href {https://doi.org/10.18653/v1/2022.naacl-main.187} {Qafacteval: Improved qa-based factual consistency evaluation for summarization}.
\newblock In \emph{Proceedings of the 2022 Conference of the North American Chapter of the Association for Computational Linguistics: Human Language Technologies, {NAACL} 2022, Seattle, WA, United States, July 10-15, 2022}, pages 2587--2601. Association for Computational Linguistics.

\bibitem[{Ghalandari et~al.(2020)Ghalandari, Hokamp, Pham, Glover, and Ifrim}]{DBLP:conf/acl/wcep20}
Demian~Gholipour Ghalandari, Chris Hokamp, Nghia~The Pham, John Glover, and Georgiana Ifrim. 2020.
\newblock \href {https://doi.org/10.18653/v1/2020.acl-main.120} {A large-scale multi-document summarization dataset from the wikipedia current events portal}.
\newblock In \emph{Proceedings of the 58th Annual Meeting of the Association for Computational Linguistics, {ACL} 2020, Online, July 5-10, 2020}, pages 1302--1308. Association for Computational Linguistics.

\bibitem[{Gliwa et~al.(2019)Gliwa, Mochol, Biesek, and Wawer}]{gliwa-etal-2019-samsum}
Bogdan Gliwa, Iwona Mochol, Maciej Biesek, and Aleksander Wawer. 2019.
\newblock \href {https://doi.org/10.18653/v1/D19-5409} {{SAMS}um corpus: A human-annotated dialogue dataset for abstractive summarization}.
\newblock In \emph{Proceedings of the 2nd Workshop on New Frontiers in Summarization}, pages 70--79, Hong Kong, China. Association for Computational Linguistics.

\bibitem[{Goyal and Durrett(2021)}]{DBLP:conf/naacl/AnnotatingFineGrained21}
Tanya Goyal and Greg Durrett. 2021.
\newblock \href {https://doi.org/10.18653/v1/2021.naacl-main.114} {Annotating and modeling fine-grained factuality in summarization}.
\newblock In \emph{Proceedings of the 2021 Conference of the North American Chapter of the Association for Computational Linguistics: Human Language Technologies, {NAACL-HLT} 2021, Online, June 6-11, 2021}, pages 1449--1462. Association for Computational Linguistics.

\bibitem[{Goyal et~al.(2022)Goyal, Li, and Durrett}]{DBLP:journals/corr/abs-2209-12356}
Tanya Goyal, Junyi~Jessy Li, and Greg Durrett. 2022.
\newblock \href {https://doi.org/10.48550/arXiv.2209.12356} {News summarization and evaluation in the era of {GPT-3}}.
\newblock \emph{CoRR}, abs/2209.12356.

\bibitem[{Grangier and Iter(2022)}]{DBLP:conf/acl/GrangierI22}
David Grangier and Dan Iter. 2022.
\newblock \href {https://doi.org/10.18653/v1/2022.acl-long.264} {The trade-offs of domain adaptation for neural language models}.
\newblock In \emph{Proceedings of the 60th Annual Meeting of the Association for Computational Linguistics (Volume 1: Long Papers), {ACL} 2022, Dublin, Ireland, May 22-27, 2022}, pages 3802--3813. Association for Computational Linguistics.

\bibitem[{Grusky et~al.(2018)Grusky, Naaman, and Artzi}]{DBLP:conf/naacl/GruskyNA18}
Max Grusky, Mor Naaman, and Yoav Artzi. 2018.
\newblock \href {https://doi.org/10.18653/v1/n18-1065} {Newsroom: {A} dataset of 1.3 million summaries with diverse extractive strategies}.
\newblock In \emph{Proceedings of the 2018 Conference of the North American Chapter of the Association for Computational Linguistics: Human Language Technologies, {NAACL-HLT} 2018, New Orleans, Louisiana, USA, June 1-6, 2018, Volume 1 (Long Papers)}, pages 708--719. Association for Computational Linguistics.

\bibitem[{Guo and Yu(2022)}]{DBLP:journals/corr/abs-2211-03154}
Xu~Guo and Han Yu. 2022.
\newblock \href {https://doi.org/10.48550/arXiv.2211.03154} {On the domain adaptation and generalization of pretrained language models: {A} survey}.
\newblock \emph{CoRR}, abs/2211.03154.

\bibitem[{Hermann et~al.(2015)Hermann, Kocisk{\'{y}}, Grefenstette, Espeholt, Kay, Suleyman, and Blunsom}]{DBLP:conf/nips/HermannKGEKSB15}
Karl~Moritz Hermann, Tom{\'{a}}s Kocisk{\'{y}}, Edward Grefenstette, Lasse Espeholt, Will Kay, Mustafa Suleyman, and Phil Blunsom. 2015.
\newblock \href {https://proceedings.neurips.cc/paper/2015/hash/afdec7005cc9f14302cd0474fd0f3c96-Abstract.html} {Teaching machines to read and comprehend}.
\newblock In \emph{Advances in Neural Information Processing Systems 28: Annual Conference on Neural Information Processing Systems 2015, December 7-12, 2015, Montreal, Quebec, Canada}, pages 1693--1701.

\bibitem[{Huang et~al.(2020)Huang, Cui, Yang, Bao, Wang, Xie, and Zhang}]{DBLP:conf/emnlp/HuangCYBWXZ20}
Dandan Huang, Leyang Cui, Sen Yang, Guangsheng Bao, Kun Wang, Jun Xie, and Yue Zhang. 2020.
\newblock \href {https://doi.org/10.18653/v1/2020.emnlp-main.33} {What have we achieved on text summarization?}
\newblock In \emph{Proceedings of the 2020 Conference on Empirical Methods in Natural Language Processing, {EMNLP} 2020, Online, November 16-20, 2020}, pages 446--469. Association for Computational Linguistics.

\bibitem[{Kornilova and Eidelman(2019)}]{kornilova-eidelman-2019-billsum}
Anastassia Kornilova and Vladimir Eidelman. 2019.
\newblock \href {https://doi.org/10.18653/v1/D19-5406} {{B}ill{S}um: A corpus for automatic summarization of {US} legislation}.
\newblock In \emph{Proceedings of the 2nd Workshop on New Frontiers in Summarization}, pages 48--56, Hong Kong, China. Association for Computational Linguistics.

\bibitem[{Kryscinski et~al.(2020)Kryscinski, McCann, Xiong, and Socher}]{DBLP:conf/emnlp/KryscinskiMXS20}
Wojciech Kryscinski, Bryan McCann, Caiming Xiong, and Richard Socher. 2020.
\newblock \href {https://doi.org/10.18653/v1/2020.emnlp-main.750} {Evaluating the factual consistency of abstractive text summarization}.
\newblock In \emph{Proceedings of the 2020 Conference on Empirical Methods in Natural Language Processing, {EMNLP} 2020, Online, November 16-20, 2020}, pages 9332--9346. Association for Computational Linguistics.

\bibitem[{Lazaridou et~al.(2021)Lazaridou, Kuncoro, Gribovskaya, Agrawal, Liska, Terzi, Gimenez, de~Masson~d'Autume, Kocisky, Ruder et~al.}]{lazaridou2021mind}
Angeliki Lazaridou, Adhi Kuncoro, Elena Gribovskaya, Devang Agrawal, Adam Liska, Tayfun Terzi, Mai Gimenez, Cyprien de~Masson~d'Autume, Tomas Kocisky, Sebastian Ruder, et~al. 2021.
\newblock Mind the gap: Assessing temporal generalization in neural language models.
\newblock \emph{Advances in Neural Information Processing Systems}, 34:29348--29363.

\bibitem[{Lewis et~al.(2020)Lewis, Liu, Goyal, Ghazvininejad, Mohamed, Levy, Stoyanov, and Zettlemoyer}]{DBLP:conf/acl/LewisLGGMLSZ20}
Mike Lewis, Yinhan Liu, Naman Goyal, Marjan Ghazvininejad, Abdelrahman Mohamed, Omer Levy, Veselin Stoyanov, and Luke Zettlemoyer. 2020.
\newblock \href {https://doi.org/10.18653/v1/2020.acl-main.703} {{BART:} denoising sequence-to-sequence pre-training for natural language generation, translation, and comprehension}.
\newblock In \emph{Proceedings of the 58th Annual Meeting of the Association for Computational Linguistics, {ACL} 2020, Online, July 5-10, 2020}, pages 7871--7880. Association for Computational Linguistics.

\bibitem[{Li et~al.(2021)Li, Ma, Yu, Wu, Gao, Ji, and McKeown}]{li2021timeline}
Manling Li, Tengfei Ma, Mo~Yu, Lingfei Wu, Tian Gao, Heng Ji, and Kathleen McKeown. 2021.
\newblock Timeline summarization based on event graph compression via time-aware optimal transport.
\newblock In \emph{Proceedings of the 2021 Conference on Empirical Methods in Natural Language Processing}, pages 6443--6456.

\bibitem[{Longpre et~al.(2021)Longpre, Perisetla, Chen, Ramesh, DuBois, and Singh}]{DBLP:conf/emnlp/LongprePCRD021}
Shayne Longpre, Kartik Perisetla, Anthony Chen, Nikhil Ramesh, Chris DuBois, and Sameer Singh. 2021.
\newblock \href {https://doi.org/10.18653/v1/2021.emnlp-main.565} {Entity-based knowledge conflicts in question answering}.
\newblock In \emph{Proceedings of the 2021 Conference on Empirical Methods in Natural Language Processing, {EMNLP} 2021, Virtual Event / Punta Cana, Dominican Republic, 7-11 November, 2021}, pages 7052--7063. Association for Computational Linguistics.

\bibitem[{Loureiro et~al.(2022)Loureiro, Barbieri, Neves, Anke, and Camacho{-}Collados}]{DBLP:conf/acl/LoureiroBNAC22}
Daniel Loureiro, Francesco Barbieri, Leonardo Neves, Luis~Espinosa Anke, and Jos{\'{e}} Camacho{-}Collados. 2022.
\newblock \href {https://doi.org/10.18653/v1/2022.acl-demo.25} {Timelms: Diachronic language models from twitter}.
\newblock In \emph{Proceedings of the 60th Annual Meeting of the Association for Computational Linguistics, {ACL} 2022 - System Demonstrations, Dublin, Ireland, May 22-27, 2022}, pages 251--260. Association for Computational Linguistics.

\bibitem[{Lu et~al.(2020)Lu, Dong, and Charlin}]{DBLP:conf/emnlp/LuDC20}
Yao Lu, Yue Dong, and Laurent Charlin. 2020.
\newblock \href {https://doi.org/10.18653/v1/2020.emnlp-main.648} {Multi-xscience: {A} large-scale dataset for extreme multi-document summarization of scientific articles}.
\newblock In \emph{Proceedings of the 2020 Conference on Empirical Methods in Natural Language Processing, {EMNLP} 2020, Online, November 16-20, 2020}, pages 8068--8074. Association for Computational Linguistics.

\bibitem[{Luu et~al.(2022)Luu, Khashabi, Gururangan, Mandyam, and Smith}]{DBLP:conf/naacl/LuuKGMS22}
Kelvin Luu, Daniel Khashabi, Suchin Gururangan, Karishma Mandyam, and Noah~A. Smith. 2022.
\newblock \href {https://doi.org/10.18653/v1/2022.naacl-main.435} {Time waits for no one! analysis and challenges of temporal misalignment}.
\newblock In \emph{Proceedings of the 2022 Conference of the North American Chapter of the Association for Computational Linguistics: Human Language Technologies, {NAACL} 2022, Seattle, WA, United States, July 10-15, 2022}, pages 5944--5958. Association for Computational Linguistics.

\bibitem[{Maynez et~al.(2020)Maynez, Narayan, Bohnet, and McDonald}]{DBLP:conf/acl/MaynezNBM20}
Joshua Maynez, Shashi Narayan, Bernd Bohnet, and Ryan~T. McDonald. 2020.
\newblock \href {https://doi.org/10.18653/v1/2020.acl-main.173} {On faithfulness and factuality in abstractive summarization}.
\newblock In \emph{Proceedings of the 58th Annual Meeting of the Association for Computational Linguistics, {ACL} 2020, Online, July 5-10, 2020}, pages 1906--1919. Association for Computational Linguistics.

\bibitem[{Myers and Sirois(2004)}]{myers2004spearman}
Leann Myers and Maria~J Sirois. 2004.
\newblock Spearman correlation coefficients, differences between.
\newblock \emph{Encyclopedia of statistical sciences}, 12.

\bibitem[{Nakayama et~al.(2018)Nakayama, Kubo, Kamura, Taniguchi, and Liang}]{doccano}
Hiroki Nakayama, Takahiro Kubo, Junya Kamura, Yasufumi Taniguchi, and Xu~Liang. 2018.
\newblock \href {https://github.com/doccano/doccano} {{doccano}: Text annotation tool for human}.
\newblock Software available from https://github.com/doccano/doccano.

\bibitem[{Nallapati et~al.(2016)Nallapati, Zhou, dos Santos, G{\"{u}}l{\c{c}}ehre, and Xiang}]{DBLP:conf/conll/NallapatiZSGX16}
Ramesh Nallapati, Bowen Zhou, C{\'{\i}}cero~Nogueira dos Santos, {\c{C}}aglar G{\"{u}}l{\c{c}}ehre, and Bing Xiang. 2016.
\newblock \href {https://doi.org/10.18653/v1/k16-1028} {Abstractive text summarization using sequence-to-sequence rnns and beyond}.
\newblock In \emph{Proceedings of the 20th {SIGNLL} Conference on Computational Natural Language Learning, CoNLL 2016, Berlin, Germany, August 11-12, 2016}, pages 280--290. {ACL}.

\bibitem[{Narayan et~al.(2018)Narayan, Cohen, and Lapata}]{DBLP:conf/emnlp/NarayanCL18}
Shashi Narayan, Shay~B. Cohen, and Mirella Lapata. 2018.
\newblock \href {https://doi.org/10.18653/v1/d18-1206} {Don't give me the details, just the summary! topic-aware convolutional neural networks for extreme summarization}.
\newblock In \emph{Proceedings of the 2018 Conference on Empirical Methods in Natural Language Processing, Brussels, Belgium, October 31 - November 4, 2018}, pages 1797--1807. Association for Computational Linguistics.

\bibitem[{Pagnoni et~al.(2021)Pagnoni, Balachandran, and Tsvetkov}]{pagnoni-etal-2021-understanding}
Artidoro Pagnoni, Vidhisha Balachandran, and Yulia Tsvetkov. 2021.
\newblock \href {https://doi.org/10.18653/v1/2021.naacl-main.383} {Understanding factuality in abstractive summarization with {FRANK}: A benchmark for factuality metrics}.
\newblock In \emph{Proceedings of the 2021 Conference of the North American Chapter of the Association for Computational Linguistics: Human Language Technologies}, pages 4812--4829, Online. Association for Computational Linguistics.

\bibitem[{Raffel et~al.(2020)Raffel, Shazeer, Roberts, Lee, Narang, Matena, Zhou, Li, and Liu}]{DBLP:journals/jmlr/RaffelSRLNMZLL20}
Colin Raffel, Noam Shazeer, Adam Roberts, Katherine Lee, Sharan Narang, Michael Matena, Yanqi Zhou, Wei Li, and Peter~J. Liu. 2020.
\newblock \href {http://jmlr.org/papers/v21/20-074.html} {Exploring the limits of transfer learning with a unified text-to-text transformer}.
\newblock \emph{J. Mach. Learn. Res.}, 21:140:1--140:67.

\bibitem[{Reddy et~al.(2023)Reddy, Elfardy, Chan, Small, and Ji}]{DBLP:journals/corr/sumren22}
Revanth~Gangi Reddy, Heba Elfardy, Hou~Pong Chan, Kevin Small, and Heng Ji. 2023.
\newblock \href {https://doi.org/10.1609/aaai.v37i11.26506} {Sumren: Summarizing reported speech about events in news}.
\newblock In \emph{Thirty-Seventh {AAAI} Conference on Artificial Intelligence, {AAAI} 2023, Thirty-Fifth Conference on Innovative Applications of Artificial Intelligence, {IAAI} 2023, Thirteenth Symposium on Educational Advances in Artificial Intelligence, {EAAI} 2023, Washington, DC, USA, February 7-14, 2023}, pages 12808--12817. {AAAI} Press.

\bibitem[{Rosin et~al.(2022)Rosin, Guy, and Radinsky}]{DBLP:conf/wsdm/RosinGR22}
Guy~D. Rosin, Ido Guy, and Kira Radinsky. 2022.
\newblock \href {https://doi.org/10.1145/3488560.3498529} {Time masking for temporal language models}.
\newblock In \emph{{WSDM} '22: The Fifteenth {ACM} International Conference on Web Search and Data Mining, Virtual Event / Tempe, AZ, USA, February 21 - 25, 2022}, pages 833--841. {ACM}.

\bibitem[{R{\"{o}}ttger and Pierrehumbert(2021)}]{DBLP:conf/emnlp/RottgerP21}
Paul R{\"{o}}ttger and Janet~B. Pierrehumbert. 2021.
\newblock \href {https://doi.org/10.18653/v1/2021.findings-emnlp.206} {Temporal adaptation of {BERT} and performance on downstream document classification: Insights from social media}.
\newblock In \emph{Findings of the Association for Computational Linguistics: {EMNLP} 2021, Virtual Event / Punta Cana, Dominican Republic, 16-20 November, 2021}, pages 2400--2412. Association for Computational Linguistics.

\bibitem[{Tejaswin et~al.(2021)Tejaswin, Naik, and Liu}]{tejaswin-etal-2021-well}
Priyam Tejaswin, Dhruv Naik, and Pengfei Liu. 2021.
\newblock \href {https://doi.org/10.18653/v1/2021.findings-acl.303} {How well do you know your summarization datasets?}
\newblock In \emph{Findings of the Association for Computational Linguistics: ACL-IJCNLP 2021}, pages 3436--3449, Online. Association for Computational Linguistics.

\bibitem[{Vaswani et~al.(2017)Vaswani, Shazeer, Parmar, Uszkoreit, Jones, Gomez, Kaiser, and Polosukhin}]{vaswani2017attention}
Ashish Vaswani, Noam Shazeer, Niki Parmar, Jakob Uszkoreit, Llion Jones, Aidan~N Gomez, {\L}ukasz Kaiser, and Illia Polosukhin. 2017.
\newblock Attention is all you need.
\newblock \emph{Advances in neural information processing systems}, 30.

\bibitem[{Vrandecic and Kr{\"{o}}tzsch(2014)}]{DBLP:journals/cacm/VrandecicK14}
Denny Vrandecic and Markus Kr{\"{o}}tzsch. 2014.
\newblock \href {https://doi.org/10.1145/2629489} {Wikidata: a free collaborative knowledgebase}.
\newblock \emph{Commun. {ACM}}, 57(10):78--85.

\bibitem[{Wang et~al.(2020)Wang, Cho, and Lewis}]{DBLP:conf/acl/WangCL20}
Alex Wang, Kyunghyun Cho, and Mike Lewis. 2020.
\newblock \href {https://doi.org/10.18653/v1/2020.acl-main.450} {Asking and answering questions to evaluate the factual consistency of summaries}.
\newblock In \emph{Proceedings of the 58th Annual Meeting of the Association for Computational Linguistics, {ACL} 2020, Online, July 5-10, 2020}, pages 5008--5020. Association for Computational Linguistics.

\bibitem[{Wolf et~al.(2019)Wolf, Debut, Sanh, Chaumond, Delangue, Moi, Cistac, Rault, Louf, Funtowicz, and Brew}]{DBLP:journals/corr/abs-1910-03771}
Thomas Wolf, Lysandre Debut, Victor Sanh, Julien Chaumond, Clement Delangue, Anthony Moi, Pierric Cistac, Tim Rault, R{\'{e}}mi Louf, Morgan Funtowicz, and Jamie Brew. 2019.
\newblock \href {http://arxiv.org/abs/1910.03771} {Huggingface's transformers: State-of-the-art natural language processing}.
\newblock \emph{CoRR}, abs/1910.03771.

\bibitem[{Xiao and Carenini(2022)}]{DBLP:journals/corr/abs-2209-03479}
Wen Xiao and Giuseppe Carenini. 2022.
\newblock \href {https://doi.org/10.48550/arXiv.2209.03479} {Entity-based spancopy for abstractive summarization to improve the factual consistency}.
\newblock \emph{CoRR}, abs/2209.03479.

\bibitem[{Yoon et~al.(2023)Yoon, Chan, and Han}]{DBLP:conf/www/pdsum23}
Susik Yoon, Hou~Pong Chan, and Jiawei Han. 2023.
\newblock \href {https://doi.org/10.1145/3543507.3583371} {Pdsum: Prototype-driven continuous summarization of evolving multi-document sets stream}.
\newblock In \emph{Proceedings of the {ACM} Web Conference 2023, {WWW} 2023, Austin, TX, USA, 30 April 2023 - 4 May 2023}, pages 1650--1661. {ACM}.

\bibitem[{Zhang et~al.(2020)Zhang, Zhao, Saleh, and Liu}]{DBLP:conf/icml/ZhangZSL20}
Jingqing Zhang, Yao Zhao, Mohammad Saleh, and Peter~J. Liu. 2020.
\newblock \href {http://proceedings.mlr.press/v119/zhang20ae.html} {{PEGASUS:} pre-training with extracted gap-sentences for abstractive summarization}.
\newblock In \emph{Proceedings of the 37th International Conference on Machine Learning, {ICML} 2020, 13-18 July 2020, Virtual Event}, volume 119 of \emph{Proceedings of Machine Learning Research}, pages 11328--11339. {PMLR}.

\bibitem[{Zhu et~al.(2021)Zhu, Liu, Mei, and Zeng}]{DBLP:conf/naacl/ZhuLMZ21}
Chenguang Zhu, Yang Liu, Jie Mei, and Michael Zeng. 2021.
\newblock \href {https://doi.org/10.18653/v1/2021.naacl-main.474} {Mediasum: {A} large-scale media interview dataset for dialogue summarization}.
\newblock In \emph{Proceedings of the 2021 Conference of the North American Chapter of the Association for Computational Linguistics: Human Language Technologies, {NAACL-HLT} 2021, Online, June 6-11, 2021}, pages 5927--5934. Association for Computational Linguistics.

\end{thebibliography}

\appendix


\section{Human Evaluation Instruction}
\label{sec:appendix-human-evaluation}
We present the instruction of our human evaluation experiments in Figure \ref{fig:human-annotation-instruction}.

\begin{figure*}[ht]
    \begin{tabular}{p{16cm}}
    \toprule
    In this study, we investigate the faithfulness performance of summarization models for data from different temporal periods. \\
    You are required to identify the faithfulness errors in the generated summaries by using the error typologies described below. \\
   
    \\
    \textbf{Factual} A summary contains a text span that is not covered by the source input but is consistent with world knowledge at the time of the article being published. \\ 
    \textbf{Outdated} A summary contains a text span that is not covered by the source input but is consistent with world knowledge before the time the article is published. \\ 
    \textbf{Non-verifiable} A summary contains a text span that is not covered by the source input and cannot be verified by any time of world knowledge. \\ 
    
    \\

    \\
    Please follow the instruction shown below to identify the faithfulness of the generated summaries. \\
    
    \textbf{Step 1}  Read the source document and the generated summary carefully.  \\
    \textbf{Step 2}  Check if all the information within the summaries can be directly entailed by the source article. If yes, mark the sample as \textit{faithful}. If no, go to the \textbf{Step 3}. \\
    \textbf{Step 3}  If any summary contains entities that are not entailed by the source. Please verify it by using Wikipedia with the knowledge of when the article was published. If it is consistent with the knowledge of the time the article was published, mark this entity as \textit{factual}. If not, but is consistent with the knowledge before the article was published, mark this entity as \textit{outdated}. If this entity cannot be verified by any time of knowledge, mark this entity as \textit{non-verifiable}.\\
    \textbf{Step 4}  If the summary is not sensical or ungrammatical, please mark the sample as \textit{non-verifiable}. \\
    
    \bottomrule
    \end{tabular}
    \vspace{-0.2cm}
   \caption{ Our human annotation instruction.
   }
    \label{fig:human-annotation-instruction}
    \vspace{-0.1cm}
\end{figure*}

\section{Automatic Evaluation Results}
\label{sess:appendix-automatic-evaluation}

Table~\ref{tab:auto-eval} shows the results of FactCC and QAFactEval scores obtained by different models.
We observe that CLIFF outperforms all other methods on all the test sets in terms of QAFactEval scores. 
On the other hand, ENT attains the best FactCC scores on the BBC datasets, while CLIFF achieves the highest FactCC scores on the CNN dataset. 

\section{Scientific Artifacts}

Our proposed \textsc{TempoSum} benchmark consists of two English news summarization datasets. All the data is collected from the Internet Archive\footnote{\href{https://archive.org/}{https://archive.org/}}. The licenses of the original news sources are applied. 

\section{Implementation Details}
\label{sess:implementation-details}

In this section, we describe the checkpoints that we use for our experiments.

\paragraph{PEGASUS:} We use ``google/pegasus-large-xsum''  and ``google/pegasus-large-cnn\_dailymail'' checkpoints provided by \citet{DBLP:journals/corr/abs-1910-03771} in our experiments. 

\paragraph{CLIFF:} 
We use the Pegasus+CLIFF checkpoints from the official code repository of CLIFF~\cite{DBLP:conf/emnlp/Cao021}. 

\paragraph{ENT:} 
We use the source codes provided by the authors to fine-tune PEGASUS. 
The authors find that the proposed mechanism is performs better on a filtered dataset (i.e., all the entities in the reference summaries are present in the source article). 
We follow the authors by using the filtered version of XSum and CNN/DM to fine-tune PEGASUS.

\paragraph{Transformer:} We use the transformer-base model from Huggingface~\cite{DBLP:journals/corr/abs-1910-03771}.

\begin{table}[t!]
\centering \footnotesize
\begin{tabular}{lcccc}
\toprule
\textbf{Data}  & \textbf{Test set} & \textbf{Model} & \textbf{FactCC} & \textbf{QAFactEval} \\
\midrule
 \multirow{8}{*}{BBC} & \multirow{4}{*}{In-dist.} & PEGASUS & 14.44 & 27.99 \\
 &  & Trans. & 17.97 & 3.99\\
 &  & CLIFF & 15.40 & \textbf{30.38} \\
 &  & ENT & \textbf{17.99} & 20.16 \\
 \cmidrule{2-5}
 & \multirow{4}{*}{Future} & PEGASUS & 11.24 & 24.97\\
  &  & Trans. & 15.45 & 2.28 \\
 &  & CLIFF & 12.04 & \textbf{29.12} \\
 &  & ENT & \textbf{16.86} & 18.87\\
 \midrule
 \multirow{8}{*}{CNN} & \multirow{4}{*}{In-dist.} & PEGASUS & 50.57 & 73.86\\
 &  & Trans. & 51.70 & 71.69 \\
 &  & CLIFF & \textbf{57.11} & \textbf{78.41} \\
 &  & ENT & 33.02 & 61.39 \\
 \cmidrule{2-5}
 & \multirow{4}{*}{Future} & PEGASUS & 61.64 & 77.64\\
  &  & Trans. & 54.54 & 73.44 \\
 &  & CLIFF & \textbf{63.42} & \textbf{80.44}\\
 &  & ENT & 35.83 & 65.23 \\
\bottomrule
\end{tabular}
\caption{
FactCC and QAFactEval scores obtained by different summarization systems on the \textsc{TempoSum} benchmark. 
}
\label{tab:auto-eval}
\end{table}

\end{document}